\documentclass[letterpaper]{article} 
\usepackage{aaai2027}  
\nocopyright 
\usepackage[hyphens]{url}  
\usepackage{graphicx} 
\usepackage{natbib}  
\usepackage{caption} 
\usepackage{booktabs}
\usepackage{amsmath}
\usepackage{amssymb}
\usepackage{amsthm}

\newtheorem{proposition}{Proposition}
\theoremstyle{remark}
\newtheorem{remark}{Remark}

\title{Policy Invariance Is Not Learning Invariance:\\Potential-Based Reward Shaping for Navigation with a Limited View}
\author{
    Nodirjon Khabibullaev\textsuperscript{\rm 1},
    Rajat Dandekar\textsuperscript{\rm 2}
}
\affiliations{
    \textsuperscript{\rm 1}Central Asian University, Tashkent, Uzbekistan\\
    \textsuperscript{\rm 2}Vizuara AI Labs\\
    khabibullaevnodirjon@gmail.com, rajatdandekar@vizuara.com
}

\begin{document}

\maketitle

\begin{abstract}
Potential-based reward shaping (PBRS) does not change the optimal policy, but how much it speeds up learning depends on the potential and on effects outside this guarantee. We analyse how PBRS changes the one-step reward while it preserves the discounted return: a per-step $(\gamma-1)\Phi$ term, the terminal-state rule, a constant shift of the potential, and positive cycles in a bonus that is not potential-based. We then measure these consequences in pre-registered experiments with a PPO agent. The agent has a limited egocentric view and is tested on two MiniGrid navigation tasks (160 runs, 10 seeds per condition). On LavaCrossing, a geodesic (BFS) potential learned faster than both the sparse baseline and a Euclidean potential with a 7$\times$7 and a 5$\times$5 view (AUC difference $+0.27$ to $+0.39$, Holm $p$ between $0.007$ and $0.0499$). The Euclidean potential did not differ from the baseline. In 20 runs with a non-potential bonus, none of the agents learned the task. The Euclidean potential caused more than twice as many deaths in lava during early training. View size did not detectably change the benefit of shaping. With a 5$\times$5 view on SimpleCrossing, some BFS runs got stuck in loops and never learned; shifting the potential did not remove this, so the per-step term alone does not explain it.
\end{abstract}

\section{Introduction}
\label{sec:intro}

In recent years, reinforcement learning (RL) has been widely used in robot navigation, computer games, and autonomous systems \citep{zhu2021deep,mnih2015human,kiran2021deep,tang2024deep}. This is because RL allows an agent to learn complex behaviours on its own by interacting with the environment. In many navigation tasks, agents get rewards only when they reach the goal. Also, agents cannot always see the whole environment. They can only see a small area around them and must find the right path while avoiding obstacles and dangerous areas. This creates a basic issue: to speed up learning, an additional guiding signal is needed, but the agent can be distracted from the original task if the signal is chosen badly \citep{ng1999policy}.

A simple way to address the sparse-reward problem is to give the agent an extra bonus for every step towards the goal. But such a bonus can change the original task. As a result, the agent may learn to circle in place and collect the bonus instead of reaching the goal. \citet{ng1999policy} described this with the bicycle-riding example.

Some methods use special rules for each task to control extra rewards \citep{zhang2020bebold}. This makes them difficult to use in different environments. Potential-based reward shaping (PBRS) was introduced as another option. It gives extra rewards based on the difference between the potential values of two states. It has been proven that PBRS does not change the optimal policy \citep{ng1999policy,wiewiora2003potential}. It has also helped reduce the negative effects of extra rewards in environments such as MiniGrid \citep{chevalier2023minigrid,forbes2024potential}. Still, PBRS only keeps the optimal policy unchanged; it does not always help the agent learn faster, because the learning speed depends on how the potential function is chosen.

In navigation tasks, the potential is usually defined by the straight-line distance to the goal (Euclidean distance) or another type of distance that does not consider walls. This is because these distances are easy to calculate and do not require a map of the walls \citep[e.g.,][]{laidlaw2023bridging}. In the early experiments on PBRS, the Manhattan distance to the goal was also used. These experiments were carried out in grid environments without walls. In such environments, the Manhattan distance is equal to the shortest-path distance \citep{ng1999policy}. However, in environments with obstacles, this distance may lead the agent towards a wall. As a result, the agent may get stuck in front of the wall and fail to find the correct open path \citep{trott2019keeping}. These problems depend on the structure of the environment. They can significantly slow down the agent's learning process \citep{trott2019keeping,veviurko2024max}. To address this problem, recent studies have proposed distances that take obstacles into account and are based on the shortest path \citep{wang2023reachability,zhang2026high}. This type of distance guides the agent along the actual path and has achieved a much higher success rate than the Euclidean distance \citep{zhang2026high}. Most of these studies, however, did not use the exact PBRS formula, and their agents could observe the full state of the environment. It is therefore still unclear how well these results apply to environments where agents can only observe their surroundings and where dangerous areas are present.

To address this research gap, we study how the choice of potential affects the learning process of PPO in a controlled experiment. The benefits or drawbacks of reward shaping depend on several factors that may change at the same time, so these factors need to be studied separately. In all experiments, the PPO settings, environments, and evaluation methods remain the same. Only the reward signal is changed. The potentials are applied using the exact PBRS formula, with a potential of zero at terminal states. This allows us to directly link the differences in results to the potential itself. The conditions, evaluation metrics, and statistical tests for each experiment were registered before the experiments began. The results are evaluated not only by the final success rate but also by the types of mistakes the agent makes. Our main contributions are:
\begin{enumerate}
\item We compare Euclidean and BFS (shortest-path) potentials, a lava-proximity term, and a non-potential bonus as a control, using the same PPO settings. All methods are tested using the exact PBRS formula with an agent that has a limited field of view. The main finding is that the BFS potential speeds up learning in the LavaCrossing environment with both 7$\times$7 and 5$\times$5 fields of view, whereas the Euclidean potential does not provide any clear benefit (Section~\ref{sec:results}).
\item We analyse how the zero potential at terminal states affects the shaped rewards when the agent reaches lava or the goal (Section~\ref{sec:analysis}). We measure this effect by looking at the agent's death rate during learning.
\item In a pre-registered test, we study how 5$\times$5 and 7$\times$7 fields of view affect the effectiveness of PBRS. The results show that the size of the field of view does not significantly change the benefits of reward shaping.
\item We show that some runs using the BFS potential with a 5$\times$5 field of view can get stuck in loops and fail to learn at all. We investigate whether this is caused by the negative potential term at each step and run an experiment with shifted potential values. Shifting the potential does not solve the problem, so the failure cannot be explained by this term alone.
\end{enumerate}

\section{Related Work}
\label{sec:related}

\paragraph{Potential-based reward shaping.}
In reinforcement learning, an agent learns only through rewards. Therefore, in tasks where the agent receives a reward only after reaching the goal, the learning process can be very slow. Reward shaping tries to solve this problem by adding an extra guiding signal to the original reward. \citet{ng1999policy} proved that the optimal policy remains unchanged if the additional reward has the following form:
\[
F(s,a,s') = \gamma\Phi(s') - \Phi(s).
\]
Here, $\Phi(s)$ is a potential function that shows how close a state is to the goal. Furthermore, without extra information about how the environment works and gives rewards, no other form of reward shaping can guarantee that the optimal policy stays the same. \citet{wiewiora2003potential} showed that PBRS is equivalent to using $\Phi$ to set the initial value function. This explains why PBRS can speed up learning: the agent does not start learning from zero but starts with an initial estimate provided by $\Phi$. How much learning is improved therefore depends on how $\Phi$ is chosen. \citet{ng1999policy} pointed out that choosing $\Phi = V^*$ is especially good, because in this case the optimal value function after reward shaping becomes zero, but that reward shaping can still be useful even when the potential function is very different from $V^*$. For episodic tasks, the potential must be zero at terminal states to keep this guarantee \citep{grzes2017reward}. This condition can also have a negative effect: if $\Phi$ has a negative value, moving to a bad terminal state, such as a dangerous state, may also give the agent an extra positive reward \citep{toroicarte2022reward}. It is important to note that keeping the optimal policy unchanged does not guarantee safety: it does not mean that the agent will avoid dangerous states during learning. According to \citet{muller2025improving}, adding a constant value to the potential function $\Phi$ does not change the optimal policy, but it does affect the learning process. This effect has not been tested with the PPO algorithm or in situations involving terminal states. PBRS does not always help in practice. For example, \citet{adamczyk2025bootstrapped} found that using the agent's own value estimates as a potential helped in some environments but made learning worse in others. \citet{laidlaw2023bridging} also showed that the effect of reward shaping depends on the structure of the task.

\paragraph{Distance potentials: Euclidean and shortest path.}
As discussed above, the effectiveness of PBRS mainly depends on how the potential function $\Phi$ is chosen. In navigation tasks, the most natural choice is the distance to the goal, but how this distance is calculated can greatly affect the results. If the goal is behind a wall, the straight-line (Euclidean) distance may guide the agent towards the wall. As a result, the agent may fail to find the correct open path and get stuck in front of the wall. In particular, \citet{trott2019keeping} showed that in continuous mazes, a reward based on the distance to the goal can lead the agent to local optima. In such tasks, the main benefit of reward shaping may not be faster learning. Instead, it may help the agent find the only available path (bottleneck) by making exploration easier. This is in line with the idea of \citet{ng1999policy} that reward shaping can help with exploration. \citet{wang2023reachability} found that in MiniGrid mazes, a special learned representation that takes obstacles into account (the reachability-aware Laplacian) gave better reward shaping results than the standard Laplacian representation. It is important to note that this was not a BFS distance but a distance learned through experience. In their experiments with PPO in mazes, \citet{veviurko2024max} showed that a reward based on the shortest-path distance worked better than a reward based on simple distance. \citet{zhang2026high} used the Dijkstra distance for drone navigation. At the highest speed (9\,m/s), the success rate was 0\% with the Euclidean distance, while it reached 43.75\% with the Dijkstra distance. However, these studies did not use the exact PBRS formula. In the study by \citet{trott2019keeping}, the reward was mainly based on the negative distance value given at the end of each episode. For this reason, it is difficult to say whether the failure of the Euclidean distance was caused by the way distance was calculated or by the way rewards were given. The comparison in Appendix~\ref{app:literature} leads to two main conclusions. First, most evidence showing that obstacle-aware distances are more useful comes from studies that did not use the exact PBRS formula. Studies that used the exact PBRS formula either used distances that did not consider walls \citep{laidlaw2023bridging} or studied tasks that were very different from ours \citep[multi-agent search, preprint]{ying2025generalizable}. Second, none of these studies directly compared Euclidean and shortest-path potentials for agents with limited vision using the same algorithm and evaluation method. In an unweighted grid environment where every step has the same cost, BFS, Dijkstra, and A$^*$ all give the same shortest-path distances, so the difference between these studies is not in the distance itself, but in the reward structure, environment, and observation conditions.

\paragraph{Partial observability and hazardous states.}
In the studies mentioned above, agents could mostly observe the full state of the environment. In practice, however, an agent's field of view is often limited. An agent in MiniGrid cannot see the entire map. It can only see a small area around itself. In our study, both the actor and the critic use only this local observation. The global map is used only during learning to calculate the potential function $\Phi$. The potential function thus adds extra information through the reward, even though the agent cannot see this information itself. The Euclidean potential is also calculated using global coordinates, so the main difference between the BFS and Euclidean potentials is whether they take obstacles into account. In this way, PBRS is similar to asymmetric actor-critic methods, where the critic receives extra information, but it is not exactly the same (Section~\ref{sec:analysis}, Remark~\ref{rem:critic}). There are different views on how effective these methods are. \citet{lambrechts2025theoretical}, as well as \citet{ebi2026informed}, showed that giving extra information to the critic can be useful. In contrast, \citet{baisero2022unbiased} showed that under partial observability, such a critic may give incorrect or biased estimates. The important difference is that in these studies, the extra information is given directly to the critic, not through the reward. In their study on safe reinforcement learning (safe RL) and shielding, \citet{carr2023safe} compared learning with and without a distance-based dense reward under both partial and full observability. They pointed out that reward shaping can have additional negative effects, especially under partial observability. Their reward, however, was not based on a potential function; they compared only two observation settings (partial and full observability) and did not separately study how reward shaping and observation conditions affect each other. In our literature review, we found no study that measured the effectiveness of PBRS by changing the size of the agent's field of view. In addition to partial observability, LavaCrossing raises another important question: dangerous terminal states. Since PBRS is based on a potential function, a negative potential near lava does not change the optimal policy or guarantee safety. Also, because $\Phi = 0$ is required at terminal states, it is not possible to directly punish the agent for the step in which it enters the lava. At the same time, since the opening is located in the same row as the lava, this potential may also punish the only available path. This effect has not yet been studied experimentally in MiniGrid.

Previous studies have shown that obstacle-aware distances work better than the Euclidean distance, but most of these studies used simple rewards and agents that could observe the full environment. In our study, we compare both potentials using the exact PBRS formula, the same PPO settings, and an agent that can only see its surroundings. We use two field-of-view sizes: 7$\times$7 and 5$\times$5. This approach helps us better understand when and why reward shaping is useful.

\section{Analysis: What PBRS Changes in One Step}
\label{sec:analysis}

PBRS turns the discounted shaping sum of an episode into a number that hardly depends on the path, so the optimal policy does not change. PPO, however, sees not the return of the whole episode but the reward of each step, short-horizon sums through GAE, and a normalised reward. The propositions below make this gap precise, and Section~\ref{sec:results} measures the consequence of each. Proofs are in Appendix~\ref{app:proofs}.

\paragraph{Notation.}
Episodic MDP with state $s$ (grid and agent pose), $\gamma \in (0,1)$, time limit $T_{\max} = 324$; the policy sees only an egocentric view of $s$. We set $\Phi = 0$ at terminal states (goal or lava); at the time limit the last state is not terminal and $\Phi$ is computed as usual. The shaping reward is
\begin{equation}
F(s,s') = \gamma\,\Phi(s') - \Phi(s), \qquad \Phi(s') := 0 \ \text{if } s' \text{ is terminal}.
\label{eq:pbrs}
\end{equation}
Potentials: Euclidean $\Phi_E = -d/D$ and BFS $\Phi_B = -b/B$, where $d$ and $b$ are the Euclidean and BFS distances from the agent to the goal and $D$ and $B$ are their values at the start, so $\Phi = -1$ at the start.

\begin{proposition}[Telescoping]
\label{prop:telescoping}
For any path $s_0, \dots, s_T$,
$\sum_{t=0}^{T-1} \gamma^t F(s_t, s_{t+1}) = \gamma^T \Phi(s_T) - \Phi(s_0)$.
\end{proposition}
If the episode ends in a terminal state, the sum equals $-\Phi(s_0)$ and does not depend on the path. At the time limit it differs from $-\Phi(s_0)$ by at most $\gamma^{324}\max|\Phi|$, with $\gamma^{324} \approx 0.04$. This is the episodic form of the guarantee of \citet{ng1999policy} \citep{grzes2017reward}: it concerns the whole episode and says nothing about a single step. That theorem is stated for policies over states, while our policy acts on egocentric observations. Proposition~\ref{prop:telescoping} carries the guarantee over, because it holds for every trajectory: the expected shaped return of any policy equals its unshaped return minus $\mathbb{E}[\Phi(s_0)]$, up to the time-limit term. The difference between the returns of any two policies, including memoryless observation-based ones, therefore changes by at most $2\gamma^{324}\max|\Phi|$.

\begin{proposition}[Per-step term]
\label{prop:perstep}
For a non-terminal $s'$ the bonus splits into two parts:
\begin{equation}
F(s,s') = \underbrace{\Phi(s') - \Phi(s)}_{\text{progress}} + \underbrace{(\gamma - 1)\,\Phi(s')}_{\text{per-step term}}.
\end{equation}
\end{proposition}
For the BFS potential the second term is $(1-\gamma)\,b(s')/B = 0.01\,b(s')/B \ge 0$. So the agent receives a positive bonus at every step even if it does not move (turning or bumping into a wall), and the further it is from the goal, the larger the bonus. For the undiscounted sum,
\begin{equation}
\sum_{t=0}^{T-1} F_t = \Phi(s_T) - \Phi(s_0) + (1-\gamma) \sum_{t=1}^{T} \bigl(-\Phi(s_t)\bigr),
\end{equation}
so for $\Phi \le 0$ the last term grows linearly with episode length. Over a 324-step timeout it reaches $3.24\,b/B$, in a 20-step successful episode about $0.1$. This does not change the discounted objective (Proposition~\ref{prop:telescoping}), but it changes the scale of the rewards seen in training and their order by episode outcome; VecNormalize normalises the reward by exactly this scale. Whether this term leads to loops is an empirical question (Section~\ref{sec:loops}).

\begin{proposition}[Shifted potential]
\label{prop:shift}
Let $\Phi' = \Phi + c$ at non-terminal states and $\Phi' = 0$ at terminal states. Then $F' = F - c(1-\gamma)$ on a non-terminal step and $F' = F - c$ on a terminal step. The discounted shaping sum of every episode that ends in a terminal state decreases by exactly $c$, and that of an episode that ends at the time limit by $c(1-\gamma^T)$.
\end{proposition}
The ranking of policies therefore changes by at most $c\gamma^{324} \approx 0.04c$, consistent with \citet{muller2025improving}. The progress part is unchanged, but for $c = 1$ (the shifted potential $\Phi = 1 - b/B$) the per-step term becomes $-0.01(1 - b/B) \le 0$: standing still becomes a cost, not a bonus. The price is that the last step into the goal also gets $F = -(1 - 1/B)$, so the goal signal is weakened.

\begin{proposition}[Terminal rule]
\label{prop:terminal}
If $\Phi \le 0$ and $\Phi = 0$ at terminal states, a step into any terminal state, lava included, receives $F = -\Phi(s) \ge 0$. In a cell $s$ next to lava this exceeds the best step towards the goal: for the BFS potential the gap is at least $\gamma\,(b(s)-1)/B > 0$ when $b(s) > 1$.
\end{proposition}
For the Euclidean potential, halfway along the path a step into lava receives about $+0.5$ and the best step towards the goal about $+0.12$ ($D = \sqrt{72} \approx 8.5$). By Proposition~\ref{prop:telescoping}, however, the shaping sum of a death episode equals that of an episode that reaches the goal, so lava is never the optimal choice. Prediction: early in training, while the policy is still close to random, shaping increases deaths (Section~\ref{sec:terminal}).

\begin{proposition}[Cycles in a non-potential bonus]
\label{prop:naive}
Let the naive bonus be $F_N(s,s') = \max(0, \Phi(s') - \Phi(s))$. For any closed cycle $s_0 \to \dots \to s_L = s_0$ on which $\Phi$ is not constant,
\begin{equation}
\sum_{t=0}^{L-1} F_N(s_t, s_{t+1}) = \tfrac12 \sum_{t=0}^{L-1} \bigl|\Phi(s_{t+1}) - \Phi(s_t)\bigr| > 0.
\end{equation}
\end{proposition}
Under PBRS the discounted sum over a cycle equals $(\gamma^L - 1)\Phi(s_0)$, which is almost zero. With the naive bonus, reward therefore accumulates as the cycle repeats: the undiscounted sum grows linearly with the number of cycles, and the discounted sum grows up to $(\tfrac12\sum|\Delta\Phi|)/(1-\gamma^L)$. Moving back and forth between two neighbouring cells ($L = 6$ actions with turns) under the Euclidean potential gives about $0.118/(1-0.99^6) \approx 2.0$, the same order as the discounted return of reaching the goal. The optimal policy can therefore change \citep{ng1999policy}.

\begin{remark}[PBRS and the critic]
\label{rem:critic}
If the critic is $V' = V - \Phi$, the shaped TD error equals the unshaped one, so GAE \citep{schulman2016gae} is also the same:
\begin{align}
\delta'_t &= r_t + F_t + \gamma V'(s_{t+1}) - V'(s_t) \nonumber \\
&= r_t + \gamma V(s_{t+1}) - V(s_t) = \delta_t.
\end{align}
\end{remark}
This is the actor-critic form of the identity $V'^{\pi} = V^{\pi} - \Phi$ \citep[Corollary~2, Remark~1]{ng1999policy} and of the equivalence of \citet{wiewiora2003potential}. It holds exactly only with $\Phi = 0$ at terminal states, correct bootstrapping at the time limit and an unnormalised reward; with VecNormalize it is approximate in our setting. The consequence is that the critic has to learn $V - \Phi$ from the egocentric observation alone, while $\Phi$ depends on the global position. A smaller view makes $\Phi$ supply information the critic lacks, but it also makes $V - \Phi$ harder to learn. Which effect dominates is the question of RQ4 (Section~\ref{sec:view}).

\section{Experimental Setup}
\label{sec:setup}

Across all conditions only the reward signal changes; the environment, the PPO configuration, the seeds and the evaluation protocol are the same. The conditions, metrics and tests of each experiment were committed before the runs (pre-registration); code, configurations and pre-registration documents are public at \url{github.com/Nodirjn/RL-Research}.

\paragraph{Questions.}
RQ1: Does a geodesic (BFS) potential learn faster than a Euclidean potential? RQ2: Does a Euclidean potential learn faster than the sparse baseline? RQ3: Does the form of the potential matter, i.e.\ what happens when the same distance signal is given as a bonus without telescoping? RQ4: Does the benefit of shaping grow when the view shrinks?

\paragraph{Environments.}
MiniGrid SimpleCrossingS9N1 (one wall line with a single opening) and LavaCrossingS9N1 (one lava line with a single opening; a step into lava ends the episode with reward 0) \citep{chevalier2023minigrid}. The grid is 9$\times$9, the start is $(1,1)$, the goal is $(7,7)$ and the time limit is 324 steps. The map is regenerated at every reset. The task reward is $r = 1 - 0.9\cdot(\text{steps}/324)$ on reaching the goal and 0 otherwise.

\paragraph{Agent.}
Stable-Baselines3 \citep{raffin2021sb3} PPO \citep{schulman2017ppo} with the MiniGrid hyperparameters of RL Zoo \citep{raffin2020zoo}, unchanged: 8 environments, \texttt{n\_steps} 128, batch 64, 10 epochs, learning rate $2.5\times10^{-4}$, $\gamma = 0.99$, $\lambda = 0.95$, clip 0.2, entropy coefficient 0, observation and reward normalisation (VecNormalize). The observation is an egocentric 7$\times$7 or 5$\times$5 view (FlatObsWrapper); the policy has no memory. Each run lasts 3M steps.

\paragraph{Conditions.}
Table~\ref{tab:conditions} lists the conditions; $b$ is the length of the shortest 4-neighbour path around walls and lava (BFS). Shaping is applied only during training; evaluation uses the task reward. Each condition and task has 10 seeds, 160 runs in total. Shifted BFS runs reuse the seeds (0--9) of the plain BFS runs.

\begin{table}[t]
\centering
\footnotesize
\setlength{\tabcolsep}{4pt}
\begin{tabular}{@{}lp{3.1cm}ll@{}}
\toprule
Condition & Bonus or potential & Tasks & View \\
\midrule
Sparse (baseline) & none & both & 7$\times$7, 5$\times$5 \\
Euclidean & $\Phi = -d/D$ & both & 7$\times$7, 5$\times$5 \\
BFS (geodesic) & $\Phi = -b/B$ & both & 7$\times$7, 5$\times$5 \\
Euclidean + lava & $\Phi_E - c$ on cells next to lava, $c = 1/D \approx 0.118$ & Lava & 7$\times$7 \\
Naive control & $F = \max(0, \Phi(s') - \Phi(s))$, Euclidean $\Phi$ & both & 7$\times$7 \\
Shifted BFS & $\Phi = 1 - b/B$ (exploratory) & Simple & 5$\times$5 \\
\bottomrule
\end{tabular}
\caption{Conditions. Every cell of the design has 10 seeds.}
\label{tab:conditions}
\end{table}

\paragraph{Evaluation.}
Every 20k steps on 20 held-out maps with the stochastic policy; at the end on 100 held-out maps with the stochastic and the greedy policy.

\paragraph{Metrics.}
Primary: area under the success curve (AUC, in $[0,1]$), final success, and steps to 50\% and 80\% success. Behaviour: greedy loop rate (share of episodes in which some state repeats), revisit ratio (share of steps that return to an already visited state), timeouts that end without crossing the line, death rate in the first 500k training steps, and the shaping sum by episode outcome.

\paragraph{Statistics.}
Pre-registered comparisons are tested on AUC with an exact two-sided permutation test (10 vs.\ 10 runs, 184{,}756 splits) and Holm correction (9 comparisons at 7$\times$7, 6 at 5$\times$5). Effect sizes are the difference of means with a 95\% bootstrap interval (10{,}000 resamples) and $P(B > A)$ \citep{agarwal2021deep}. RQ4 is tested with the shaping $\times$ view interaction contrast and a Bonferroni-corrected 98.75\% bootstrap interval. All other results are exploratory: they are reported with intervals and are not claims.

\section{Results}
\label{sec:results}

On LavaCrossing the BFS potential learned faster than both the sparse baseline and the Euclidean potential with both views, while the Euclidean potential did not differ from the baseline. None of 20 runs with the naive bonus learned. View size did not detectably change the benefit of shaping, but on SimpleCrossing with a 5$\times$5 view some BFS runs got stuck in loops, and shifting the potential did not fix this.

\begin{table*}[t]
\centering
\small
\begin{tabular}{@{}lll@{\hskip 8pt}l@{}}
\toprule
Comparison & Task & 7$\times$7 view & 5$\times$5 view \\
\midrule
BFS $-$ Euclidean & Lava & $+0.39$ [$+0.18$, $+0.60$], $p = 0.014$ & $+0.34$ [$+0.15$, $+0.54$], $p = 0.010$ \\
BFS $-$ sparse & Lava & $+0.27$ [$+0.09$, $+0.48$], $p = 0.0499$ & $+0.35$ [$+0.18$, $+0.54$], $p = 0.007$ \\
Euclidean $-$ sparse & Lava & $-0.12$ [$-0.39$, $+0.16$], $p = 1.0$ & $+0.02$ [$-0.23$, $+0.26$], $p = 0.91$ \\
Euclidean $-$ sparse & Simple & $+0.02$ [$-0.06$, $+0.09$], $p = 1.0$ & $-0.12$ [$-0.16$, $-0.07$], $p = 0.002$ \\
BFS $-$ sparse & Simple & $-0.01$ [$-0.11$, $+0.09$], $p = 1.0$ & $-0.29$ [$-0.57$, $-0.02$], $p = 0.22$ \\
BFS $-$ Euclidean & Simple & $-0.02$ [$-0.12$, $+0.06$], $p = 1.0$ & $-0.17$ [$-0.46$, $+0.09$], $p = 0.42$ \\
Naive $-$ Euclidean & Simple & $-0.90$ [$-0.94$, $-0.86$], $p < 0.001$ & --- \\
Naive $-$ Euclidean & Lava & $-0.45$ [$-0.65$, $-0.24$], $p = 0.019$ & --- \\
Lava term $-$ Euclidean & Lava & $+0.02$ [$-0.29$, $+0.32$], $p = 1.0$ & --- \\
\bottomrule
\end{tabular}
\caption{Pre-registered comparisons: AUC difference [95\% bootstrap interval] and Holm-corrected $p$ (10 vs.\ 10 runs).}
\label{tab:main}
\end{table*}

\begin{figure*}[t]
\centering
\includegraphics[width=\textwidth]{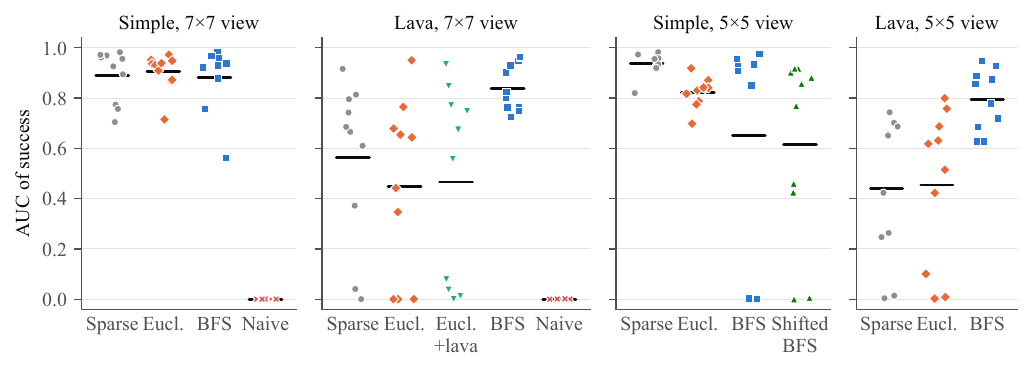}
\caption{AUC of every run (points) and the condition mean (bar); panels are task $\times$ view (Simple: SimpleCrossing, Lava: LavaCrossing). Each condition has 10 runs.}
\label{fig:auc}
\end{figure*}

\subsection{Why the Baseline Fails}
\label{sec:baseline}
\begin{figure*}[t]
\centering
\includegraphics[width=\textwidth]{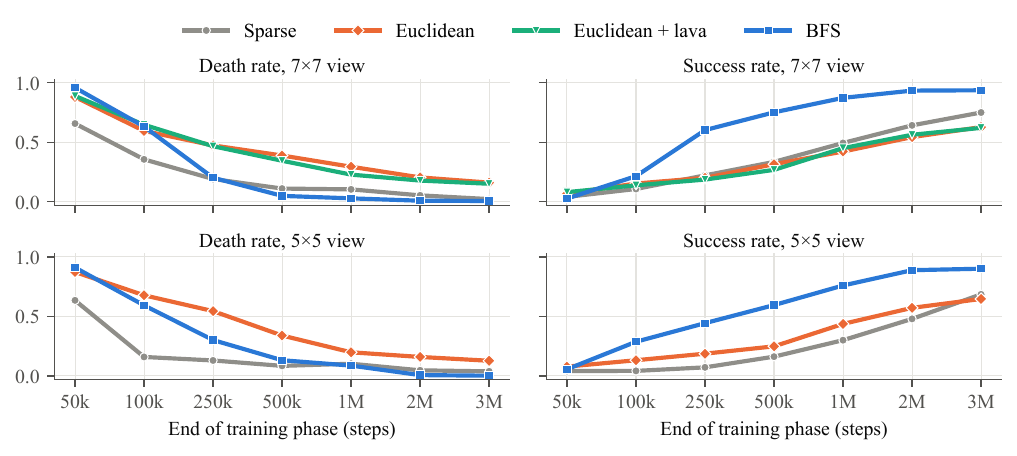}
\caption{LavaCrossing: death rate (left) and success rate (right) of training episodes by training phase; 7$\times$7 view at the top, 5$\times$5 view at the bottom. Each line is the mean of 10 runs; the remaining share is timeouts.}
\label{fig:lava}
\end{figure*}

On LavaCrossing with a 7$\times$7 view, the final stochastic episodes of sparse PPO have a death rate of 0.018 and a timeout rate of 0.222; the greedy policy never dies and times out in 0.380 of episodes (5 seeds). Deaths are frequent only early in training (61\% of episodes in the first 50k steps); after about 100k steps failures are mostly timeouts (Figure~\ref{fig:lava}). Of the timeouts, 88\% end without crossing the lava line, and every greedy timeout is a loop. The agent fails not because it dies but because it cannot find the opening: this is an exploration problem.

\subsection{Geodesic versus Euclidean Potential (RQ1, RQ2)}
\label{sec:geodesic}
Table~\ref{tab:main} shows the pre-registered comparisons and Figure~\ref{fig:auc} shows every run. The advantage of BFS is linked to passing through the opening. On LavaCrossing at 7$\times$7, timeouts without crossing the line make up 0.231 (Euclidean) and 0.061 (BFS) of all stochastic episodes, and the revisit ratio is 0.299 and 0.093. At 5$\times$5 the same share among greedy episodes is 0.386 and 0.166 (exploratory). The BFS $-$ sparse result at 7$\times$7 ($p = 0.0499$) is borderline; at 5$\times$5 it replicated more clearly. On SimpleCrossing at 7$\times$7 the baseline reaches an AUC of 0.89, so the potential has almost no room to gain.

\subsection{Terminal Rule and Lava}
\label{sec:terminal}
As Proposition~\ref{prop:terminal} predicts, shaping increased deaths in lava early in training. In training episodes of the first 500k steps, the death rate was 0.225 with sparse, 0.289 with BFS and 0.533 with Euclidean at 7$\times$7, and 0.187, 0.430 and 0.582 at 5$\times$5 (Euclidean $-$ sparse at 5$\times$5: $+0.39$ [$+0.30$, $+0.48$]; exploratory). The lava-proximity term did not change the AUC ($+0.02$, $p = 1.0$).

\subsection{Form of the Potential (RQ3)}
\label{sec:form}
None of 20 runs with the naive bonus learned (AUC 0.00). The greedy policy loops in almost every episode (loop rate 1.0 and 0.997). The undiscounted shaping sum of a timeout episode is 11.0--11.5, against 2.05--2.50 with Euclidean PBRS. The distance signal is the same as in Euclidean PBRS; the only difference is the form of the bonus: no telescoping, and positive cycles (Proposition~\ref{prop:naive}).

\subsection{View Size (RQ4)}
\label{sec:view}
The pre-registered metric is the number of steps from crossing the line to the goal (successful episodes, stochastic policy). The interaction contrast is $I = (S - \text{sparse}, 5{\times}5) - (S - \text{sparse}, 7{\times}7)$ for a shaped condition $S$, and the H-view prediction is $I < 0$. The 98.75\% intervals of all four contrasts contain 0: SimpleCrossing Euclidean $-2.2$ [$-5.5$, $+0.9$], SimpleCrossing BFS $-1.7$ [$-7.4$, $+3.1$], LavaCrossing Euclidean $-4.5$ [$-17.7$, $+6.6$] and LavaCrossing BFS $+2.7$ [$-2.5$, $+8.2$] steps. We therefore found no evidence that shaping helps more with a smaller view. The AUC interaction is reported only as an estimate: it is negative on SimpleCrossing (Euclidean $-0.13$ [$-0.22$, $-0.04$], BFS $-0.28$ [$-0.57$, $-0.01$]) and close to 0 on LavaCrossing.

\subsection{Loops with BFS at 5$\times$5 (Exploratory)}
\label{sec:loops}
\begin{figure}[!t]
\centering
\includegraphics[width=\columnwidth]{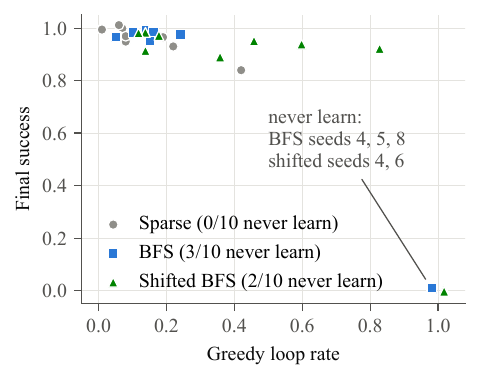}
\caption{SimpleCrossing, 5$\times$5 view: greedy loop rate (share of evaluation episodes with a loop) and final success of the stochastic policy for every run (sparse, BFS and shifted BFS; 10 runs each). Points in the lower right corner are runs that never learned: 3 with BFS and 2 with shifted BFS.}
\label{fig:loops}
\end{figure}

On SimpleCrossing at 5$\times$5, 3 of 10 BFS runs (seeds 4, 5 and 8) never learned: their AUC is 0.00, almost every training episode timed out at 324 steps, and the greedy loop rate is 1.0. The other 7 runs reached 0.94--1.00; all 10 BFS runs at 7$\times$7 learned. In the failed runs, a timeout episode collected more total reward than an episode that reached the goal (undiscounted 2.51--2.77 against 2.11--2.14), which is consistent with Proposition~\ref{prop:perstep}. To test this explanation we ran 10 runs with the shifted BFS potential (Proposition~\ref{prop:shift}, $c = 1$). The shifted potential did not remove the failure: 2 of 10 runs (seeds 4 and 6) never learned, against none with the sparse baseline (Figure~\ref{fig:loops}). The per-step term now worked in reverse: in the 8 runs that learned, timeout episodes received $-0.61$ to $-0.85$ and goal episodes $-0.10$ to $-0.16$ of undiscounted shaping. Still, the greedy loop rate did not decrease (0.47 against 0.41 with plain BFS; difference $+0.06$ [$-0.27$, $+0.37$]), the AUC did not differ from plain BFS ($-0.04$ [$-0.38$, $+0.31$]), and it was below the sparse baseline ($-0.32$ [$-0.55$, $-0.12$]). The price predicted by Proposition~\ref{prop:shift} was visible: at 500k steps the training success of 5 of 10 runs was still below 0.05 (3 with plain BFS). So the per-step term alone does not explain the loops, and the cause of this failure remains open. With this few runs the result only indicates a direction.

\section{Discussion and Limitations}
\label{sec:discussion}

Policy invariance does not imply invariance of the learning dynamics: bonuses with the same discounted objective gave very different results in training. \citet{trott2019keeping} observed the same for a distance reward that keeps the global optimum: learning settled in local optima, even in a bit-flipping task without walls. The geodesic potential helped where the failure of the baseline was an exploration problem: on LavaCrossing the agent cannot find the opening, and BFS points exactly along the path through it. On SimpleCrossing the baseline already reaches almost 100\%, so the potential has no room to gain.

Three mechanisms outside the PBRS guarantee appeared in the experiments. The terminal rule increased deaths early in training (Proposition~\ref{prop:terminal}, Section~\ref{sec:terminal}). The bonus without telescoping broke learning entirely (Proposition~\ref{prop:naive}, Section~\ref{sec:form}). The third, the loop runs at 5$\times$5, looked consistent with the per-step term of Proposition~\ref{prop:perstep}, but the shifted potential, which reverses this term, did not remove the loops either (Section~\ref{sec:loops}). The cause therefore lies elsewhere. Plausible candidates are that with a small view the critic cannot learn $V - \Phi$ (Remark~\ref{rem:critic}), or the scale of shaping under VecNormalize; we did not test these here.

The result for RQ4 is that no difference was found. Possible reasons are that the path after the line is short (6--14 steps), that the intervals are wide, and that the two effects in Remark~\ref{rem:critic} ($\Phi$ supplies extra information, but the critic finds $V - \Phi$ harder to learn) may cancel. The negative AUC interaction estimate on SimpleCrossing points towards the second effect, but this is not a claim.

\paragraph{Limitations.}
We used one PPO configuration (RL Zoo defaults) without per-condition hyperparameter tuning. Reward normalisation (VecNormalize) changes the scale of the shaping signal and makes the identity in Remark~\ref{rem:critic} approximate. There are two small tasks (9$\times$9), 10 seeds per condition, and a memoryless policy. $\Phi$ is computed from the global map and used only in training; the effects of information and of shaping are not separated by an ablation (for example, giving the BFS distance to the critic directly). Map-free alternatives that address the local optima of a distance reward, such as Sibling Rivalry \citep{trott2019keeping}, were not compared. The shifted-potential experiment was chosen after seeing the results, so it is exploratory.

\section{Conclusion}
\label{sec:conclusion}

We compared Euclidean and geodesic potentials in the exact PBRS form, a bonus that is not potential-based, and two view sizes in one pre-registered protocol. The geodesic potential sped up learning on LavaCrossing with both views, while the Euclidean potential did not. The non-potential bonus broke learning entirely, and effects outside the policy-invariance guarantee were measured as deaths and loops. View size did not detectably change the benefit of shaping. With a 5$\times$5 view, some BFS runs got stuck in loops and never learned, and shifting the potential did not fix this; the cause of this failure is an open question for future work.

\bibliography{references}

\appendix

\section{Proofs}
\label{app:proofs}

\paragraph{Proposition~\ref{prop:telescoping}.}
Shift the index by one and subtract the two sums:
\begin{align*}
\sum_{t=0}^{T-1} \gamma^t \bigl(\gamma\Phi(s_{t+1}) - \Phi(s_t)\bigr)
&= \sum_{t=1}^{T} \gamma^{t}\Phi(s_t) - \sum_{t=0}^{T-1} \gamma^t\Phi(s_t) \\
&= \gamma^T\Phi(s_T) - \Phi(s_0).
\end{align*}
At a terminal state $\Phi(s_T) = 0$. At the time limit $T = 324$ and $|\gamma^T \Phi(s_T)| \le \gamma^{324}\max|\Phi|$. \qed

\paragraph{Proposition~\ref{prop:perstep}.}
Add and subtract $\Phi(s')$: $\gamma\Phi(s') - \Phi(s) = [\Phi(s') - \Phi(s)] + (\gamma-1)\Phi(s')$. At a terminal step $\Phi(s') = 0$, so the split holds there as well. Summing over steps, the first part telescopes to $\Phi(s_T) - \Phi(s_0)$ and the second gives $-(1-\gamma)\sum_{t=1}^{T}\Phi(s_t)$. If $\Phi = -b/B$ and $b/B \le m$ on all steps, the second part is at most $0.01\,T\,m$, i.e.\ $3.24\,m$ for $T = 324$. \qed

\paragraph{Proposition~\ref{prop:shift}.}
For a non-terminal $s'$, $F' = \gamma(\Phi(s') + c) - (\Phi(s) + c) = F - c(1-\gamma)$; for a terminal $s'$, $F' = 0 - (\Phi(s) + c) = F - c$. Applying Proposition~\ref{prop:telescoping} to $\Phi'$ gives the sum $\gamma^T\Phi'(s_T) - \Phi(s_0) - c$. At a terminal state $\Phi'(s_T) = 0$, so the change is $-c$. At the time limit the sum is $\gamma^T(\Phi(s_T) + c) - \Phi(s_0) - c$, so the change is $-c(1-\gamma^T)$. The two changes differ by $c\gamma^{324}$, so the ranking of policies changes by at most this amount. For shifted BFS, $(\gamma-1)\Phi'(s') = -(1-\gamma)(1 - b(s')/B)$, and the step into the goal from $b = 1$ gives $F' = -(1 - 1/B)$. \qed

\paragraph{Proposition~\ref{prop:terminal}.}
At a terminal step $F = -\Phi(s) \ge 0$ because $\Phi \le 0$. Under the BFS potential one step decreases $b$ by at most 1, so the best forward step has $F \le (b(s) - \gamma(b(s)-1))/B$. The difference is $b(s)/B - (b(s) - \gamma(b(s)-1))/B = \gamma(b(s)-1)/B$. In the same way, under the Euclidean potential $F \le (\gamma + (1-\gamma)d(s))/D$; for $d(s) = D/2$ and $D = \sqrt{72}$ this is $0.12$, while the step into lava gives $d(s)/D = 0.5$. By Proposition~\ref{prop:telescoping}, death and goal episodes have the same discounted shaping sum ($-\Phi(s_0) = 1$), so only the task reward separates them. \qed

\paragraph{Proposition~\ref{prop:naive}.}
On a cycle $\sum_t \Delta\Phi_t = 0$, so the sum of the positive parts equals the sum of the negative parts, and each equals $\tfrac12\sum_t|\Delta\Phi_t|$. If the cycle is repeated, the discounted sum is $\sum_j \gamma^{jL}$ times the value of one cycle, which is at most $(\tfrac12\sum|\Delta\Phi|)/(1-\gamma^L)$. For a cycle between two neighbouring cells, $\tfrac12\sum|\Delta\Phi| = |\Delta\Phi| \le 1/D \approx 0.118$ and $L = 6$ (forward, two turns, forward, two turns), which gives $0.118/(1-0.99^6) \approx 2.0$. Under PBRS the cycle sum is $(\gamma^L - 1)\Phi(s_0) = (1-\gamma^L)|\Phi(s_0)|$ by Proposition~\ref{prop:telescoping}; this is the small per-step term of Proposition~\ref{prop:perstep}, at most about $0.06$ for $L = 6$. \qed

\section{Settings and Pre-Registration}
\label{app:settings}

\begin{table}[ht]
\centering
\small
\setlength{\tabcolsep}{4pt}
\begin{tabular}{@{}ll@{}}
\toprule
Parameter & Value \\
\midrule
Algorithm & PPO (Stable-Baselines3) \\
Hyperparameters & RL Zoo, MiniGrid \\
Parallel environments & 8 \\
\texttt{n\_steps}, batch, epochs & 128, 64, 10 \\
Learning rate & $2.5\times10^{-4}$ \\
$\gamma$, $\lambda$, clip & 0.99, 0.95, 0.2 \\
Entropy coefficient & 0 \\
Normalisation & VecNormalize (obs.\ and reward) \\
Observation & FlatObsWrapper; 7$\times$7 or 5$\times$5 view \\
Steps & 3M per run \\
Evaluation, training & 20 maps every 20k steps \\
Evaluation, final & 100 maps, stochastic and greedy \\
\bottomrule
\end{tabular}
\caption{PPO and evaluation settings, identical in all conditions.}
\label{tab:settings}
\end{table}

{\raggedright The pre-registration documents are in \url{documentation/stages/} of the repository: \texttt{08\_shaping\_experiment.md} (7$\times$7 view), \texttt{10\_view\_shaping\_experiment.md} (5$\times$5 view) and \texttt{12\_shifted\_potential\_experiment.md} (shifted BFS). The baseline behaviour analysis is in \texttt{05\_behavior\_analysis.md}.\par}

\section{Full Results}
\label{app:results}

\begin{table*}[t]
\centering
\small
\begin{tabular}{@{}llcccc@{}}
\toprule
Task, view & Condition & AUC & Final & Never & Death \\
 & & mean $\pm$ SD & success & learn & 500k \\
\midrule
Simple 7$\times$7 & Sparse & 0.89 $\pm$ 0.10 & 1.00 & 0 & --- \\
Simple 7$\times$7 & Euclidean & 0.90 $\pm$ 0.08 & 0.97 & 0 & --- \\
Simple 7$\times$7 & BFS & 0.88 $\pm$ 0.13 & 0.98 & 0 & --- \\
Simple 7$\times$7 & Naive & 0.00 $\pm$ 0.00 & 0.00 & 10 & --- \\
Lava 7$\times$7 & Sparse & 0.56 $\pm$ 0.32 & 0.76 & 1 & 0.225 \\
Lava 7$\times$7 & Euclidean & 0.45 $\pm$ 0.35 & 0.61 & 3 & 0.533 \\
Lava 7$\times$7 & Eucl.\ + lava & 0.47 $\pm$ 0.39 & 0.64 & 1 & 0.563 \\
Lava 7$\times$7 & BFS & 0.84 $\pm$ 0.09 & 0.93 & 0 & 0.289 \\
Lava 7$\times$7 & Naive & 0.00 $\pm$ 0.00 & 0.00 & 10 & 0.077 \\
Simple 5$\times$5 & Sparse & 0.94 $\pm$ 0.05 & 0.96 & 0 & --- \\
Simple 5$\times$5 & Euclidean & 0.82 $\pm$ 0.06 & 0.93 & 0 & --- \\
Simple 5$\times$5 & BFS & 0.65 $\pm$ 0.45 & 0.68 & 3 & --- \\
Simple 5$\times$5 & Shifted BFS & 0.61 $\pm$ 0.37 & 0.76 & 2 & --- \\
Lava 5$\times$5 & Sparse & 0.44 $\pm$ 0.29 & 0.71 & 2 & 0.187 \\
Lava 5$\times$5 & Euclidean & 0.45 $\pm$ 0.31 & 0.71 & 1 & 0.582 \\
Lava 5$\times$5 & BFS & 0.79 $\pm$ 0.12 & 0.89 & 0 & 0.430 \\
\bottomrule
\end{tabular}
\caption{Every cell has 10 runs. ``Never learn'': final success below 0.1. ``Death 500k'': share of training episodes that end in lava during the first 500k steps. Source: \texttt{results/tables/exp020\_runs.csv} and \texttt{exp022\_runs.csv} (the latter also contains the EXP-012 and EXP-021 runs).}
\label{tab:full}
\end{table*}

\section{Work on Distance-Based Shaping}
\label{app:literature}

\begin{table*}[t]
\centering
\footnotesize
\setlength{\tabcolsep}{3pt}
\begin{tabular}{@{}p{2.1cm}p{2.3cm}p{2.5cm}p{1.6cm}p{2.7cm}p{1.6cm}p{3.0cm}@{}}
\toprule
Work & Environment & Distance or potential & Exact PBRS? & Observation & Algorithm & Limitation \\
\midrule
\citet{trott2019keeping} & Continuous 2D point maze, Ant U-maze & Distance to goal, mostly at episode end (ignores walls) & No & Full state (agent and goal coordinates) & PPO & Agent falls into local optima \\
\citet{laidlaw2023bridging} & MiniGrid and other environments & Distance that ignores walls (fewest steps without obstacles) & Yes ($\gamma = 1$) & Not stated & PPO, DQN & Distance ignores obstacles \\
\citet{wang2023reachability} & Discrete gridworlds and continuous mazes & Learned representation distance & No & Agent $(x, y)$ position & DQN (discrete) & Distance is learned, not BFS \\
\citet{veviurko2024max} & Gymnasium-Robotics Maze (continuous) & Shortest-path distance (discretised) & No (max-reward RL) & State vector (position, velocity) & PPO, TD3 & Result depends on maze layout \\
\citet{zhang2026high} & Drone, continuous 3D & Dijkstra distance & No (progress reward) & Actor: depth image + proprioception; critic: exact state & PPO (asymmetric actor-critic) & Reward form and distance effect not separated \\
\citet{ying2025generalizable} & Multi-agent search, cluttered & A$^*$ distance (on agent's internal map) & Yes ($\gamma = 0.99$) & Partial: limited field of view, LiDAR, local map & MAPPO & Preprint, different task \\
This work & MiniGrid SimpleCrossing, LavaCrossing & Euclidean and BFS, lava term & Yes (terminal $\Phi = 0$) & Egocentric 7$\times$7 and 5$\times$5 & PPO & $\Phi$ from global map; no information ablation \\
\bottomrule
\end{tabular}
\caption{Work on distance-based shaping compared with this study.}
\label{tab:literature}
\end{table*}

\end{document}